\documentclass[runningheads]{llncs}

\usepackage{eccv}

\usepackage{eccvabbrv}

\usepackage{graphicx}
\usepackage{booktabs}

\usepackage[accsupp]{axessibility}  % Improves PDF readability for those with disabilities.

\usepackage[pagebackref,breaklinks,colorlinks,citecolor=eccvblue]{hyperref}

\usepackage{amsmath}
\usepackage{amssymb}
\usepackage{csquotes}
\usepackage{booktabs} 
\usepackage{array}
\usepackage{adjustbox}
\usepackage{xspace}
\usepackage{multirow}
\usepackage{wrapfig}
\usepackage{enumitem}
\usepackage{bbm}
\usepackage{makecell}
\usepackage{pifont}
\usepackage{listings}
\usepackage{xcolor}
\usepackage{colortbl}
\DeclareMathOperator*{\argmax}{arg\,max}

\newcommand{\x}{\boldmath{x}}
\newcommand{\m}{\boldmath{m}}

\newcommand{\M}{\mathcal{M}}

\newcommand{\RR}{\mathbb{R}}

\usepackage{orcidlink}
\usepackage[capbesideposition=outside,capbesidesep=quad]{floatrow}
\usepackage{sidecap}

\begin{document}

\title{Adversarially Masked Video Consistency for Unsupervised Domain Adaptation} 

\titlerunning{Adversarially Masked Video Consistency for
 Domain Adaptation}

\author{Xiaoyu Zhu\inst{1} \and
Junwei Liang\inst{2}\and Po-Yao Huang\inst{3} \and
Alex Hauptmann\inst{1}}

\authorrunning{X. Zhu et al.}

\institute{Carnegie Mellon University \and HKUST \and FAIR, Meta}

\maketitle

\begin{center}
% \vspace{-1mm}
  \includegraphics[width=1.0\linewidth, height=4.5cm]{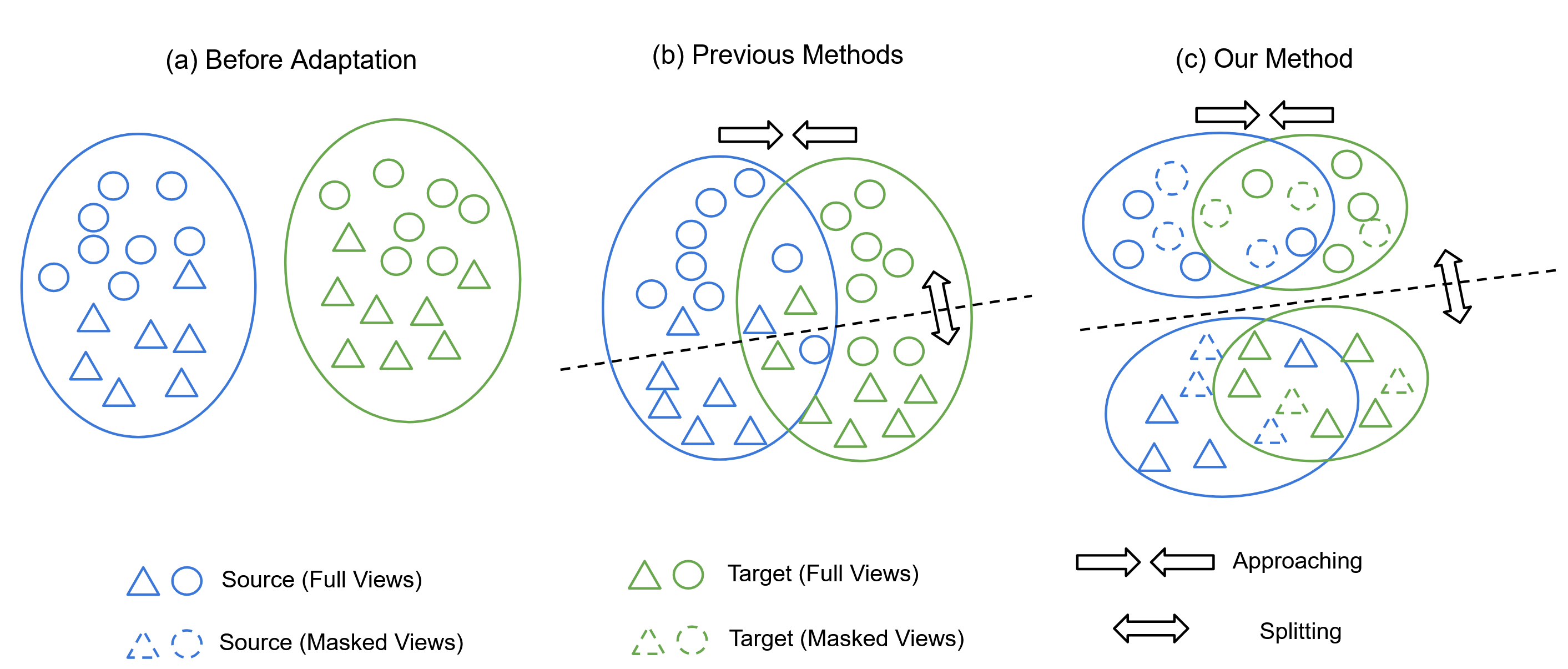}
% \vspace{-2mm}
\captionof{figure}{Visualization of the feature space for unsupervised domain adaptation methods.
Existing state-of-the-art video domain adaptation models \cite{tzeng2017adversarial,ganin2016domain,
chen2019temporal} used full-view input data to perform domain alignment as shown in (b). In this work, we propose a model 
that learns from adversarially masked samples, which can lead to the learning of effective domain-invariant and class-discriminative representations.}
\label{fig1}
\end{center}

\begin{abstract}
We study the problem of unsupervised domain adaptation for egocentric videos. We propose a transformer-based model 
to learn class-discriminative and domain-invariant feature representations. 
It consists of two novel designs. The first module is called Generative Adversarial Domain Alignment Network with the aim of learning domain-invariant representations. It simultaneously learns a mask generator
and a domain-invariant encoder in an adversarial way. The domain-invariant encoder is trained to minimize
the distance between the source and target domain. The masking
generator, conversely, aims at producing challenging masks by maximizing the domain distance.
The second is a Masked Consistency Learning module to learn class-discriminative representations. It enforces the
prediction consistency between the masked target videos and their full forms. 
To better evaluate the effectiveness of domain adaptation methods, we construct a more challenging benchmark for egocentric videos, U-Ego4D.
Our method achieves state-of-the-art 
performance on the Epic-Kitchen and the proposed U-Ego4D benchmark. 
\end{abstract}    
\section{Introduction}
Egocentric vision \cite{sudhakaran2019lsta,kazakos2019epic,munro20multi,Ego4D2022CVPR,nagarajan2020ego,liu2019forecasting,li2019deep,zhang2022fine,Liu_2022_CVPR,swan2007egocentric,jones2008effects,liang2015ar} has attracted increasing attention in the computer vision community. It serves as key
elements for various research fields, such as human-object interaction \cite{zhang2022fine,Liu_2022_CVPR}, action recognition \cite{sudhakaran2019lsta,kazakos2019epic,munro20multi}, action anticipation \cite{nagarajan2020ego,liu2019forecasting}, social interaction analysis \cite{li2019deep}, and augmented reality \cite{swan2007egocentric,jones2008effects,liang2015ar}. Besides, egocentric vision has been
popular in many real-world applications \cite{kazakos2019epic} including health monitoring, life logging, and home automation. Among egocentric perception tasks, egocentric action recognition is an important and challenging task compared to third-person action recognition, due to the
presence of ego-motion caused by the action performer. Such camera motion introduces heavy noises that complicate the extraction of visual representation from the video frames \cite{munro20multi}.
Moreover, egocentric action recognition task usually requires high-fidelity modeling of human behaviors \cite{zhang2022finegrained}, such as \emph{cut a vegetable} instead of coarse-grained actions such as \emph{cook}.
This requires the model to effectively recognize small objects and their mutual interaction. To train a discriminate model that is robust to sharp domain gaps, supervised approaches rely on collecting and annotating a large
number of videos, which is expensive and may not be feasible in practice.

To address the lack of fine-grained data annotations, Unsupervised Domain Adaptation (UDA) setting is commonly used to transfer a model learned on a labeled source domain to an unlabelled target domain. However, existing unsupervised domain adaptation benchmarks for egocentric action recognition are limited to a single environment \cite{Damen2018EPICKITCHENS,Damen2022RESCALING} (\emph{i.e.} kitchens), with small domain variances (\emph{i.e.} different kitchens are treated as different domains).
As a first step in this direction, we propose a new unsupervised domain adaptation benchmark, named U-Ego4D. We leverage the massive-scale Ego4D dataset \cite{Ego4D2022CVPR}. It records daily-life activity videos spanning hundreds of scenarios (household, outdoor, workplace, leisure, etc.).  
The proposed U-Ego4D treats actions in different regions as different domains, which is more challenging. 
Moreover, the same action can happen in the same or different scenarios. For example, the action \emph{cut}, can happen in a kitchen such as \emph{cut dough} and \emph{cut a vegetable}, or happen outdoors such as \emph{cut grass}. 

There are several challenges in training a model that is robust to various scenarios with only labeled source data available. First, there are multiple factors that could lead to domain gaps, including different backgrounds, lighting conditions, viewpoints, interacted objects, and motion variances. To bridge the domain gap, most of the existing state-of-the-art methods use adversarial learning to align two domains based on the full-view data. However, it might lead to trivial solutions (\emph{i.e.} the model might be over-fitted to differentiate the source and target domain only based on lighting differences, while other factors are neglected). Second, the decision boundary learned on labeled source videos may generalize poorly to the target domain. The model may overfit the source data well but is less discriminative for the target.

To tackle the aforementioned challenges,  we propose a transformer-based model that utilizes masked data to avoid trivial solutions and learns more generalizable representations. 
This is different from existing state-of-the-art methods, which only take full-view data as inputs, as illustrated in Figure~\ref{fig1}. Our model consists of two novel designs: Generative Adversarial Domain Alignment Network (GADAN) and a Masked Consistency Learning (MCL) module.
GADAN simultaneously learns a masking generator and a
domain-invariant encoder in an adversarial way. The domain-invariant encoder
is trained to minimize the feature distance between the source and target domain. The
mask generator, conversely, aims at producing challenging masks by maximizing
the domain distance. 
To increase the model's class-discriminative ability, MCL
enforces the prediction consistency between the masked target videos and their
full forms, and enhances the understanding of spatial-temporal context. 
We show the efficacy of our model on the Epic-Kitchen and the proposed U-Ego4D benchmarks. 
Our contributions are three-fold:
\begin{itemize}
  \item We propose the U-Ego4D benchmark, to enable the evaluation of video domain adaptation models in a more challenging and practical scenario.
  \item We introduce a new transformer-based model, which contains the Generative Adversarial Domain Alignment Network and the Masked Consistency Learning module to learn effective domain-invariant and class-discriminative representations.
  \item Our method outperforms existing state-of-the-art models on Epic-Kitchen and U-Ego4D benchmarks.
\end{itemize}

\section{Related Work}

\paragraph{Egocentric Vision.} Egocentric vision is more complicated compared to third-person perception. It brings various challenges, such as sharp viewpoint movement, object occlusions, and environmental bias~\cite{plizzari20212,li2015delving,munro2020multi, planamente2021domain, song2021spatio, kim2021learning, sahoo2021contrast}. To help the model focus on the regions of interest and better recognize different actions, Sudhakaran \cite{sudhakaran2019lsta} proposed to use both long-term and short-term attention mechanisms to recognize fine-grained actions. Lu \cite{8653357} introduced a two-stream deep neural network which consists of an appearance-based stream and a motion-based stream for action recognition. Another stream focuses on leveraging multi-modal information, such as RGB, depth, audio, and event camera \cite{Kazakos_2019_ICCV,li2021trear,plizzari2022e2}. \cite{Kazakos_2019_ICCV} introduced a novel architecture for multi-modal temporal-binding. It is able to combine multiple modalities within a range of temporal offsets. The proposed framework combined three modalities (\emph{i.e.}  RGB, Flow and Audio) for egocentric action recognition.  \cite{li2021trear} proposed a transformer-based method which includes inter-frame attention encoder and mutual-attentional fusion block. The model consumes both RGB and depth images as inputs. \cite{plizzari2022e2} proposed to use event-camera data to distill optical flow information. Leveraging event-camera data is demonstrated to be effective and can improve performance of up to 4\% with respect to RGB only information.
However, those methods need extra sensors and increase the computational cost, as multiple backbone networks are needed to encode different modalities. 
To tackle the challenges brought by egocentric videos, we propose a masked consistency learning module to help the model learn the spatial-temporal context.

\paragraph{Video Domain Adaptation.} Video domain adaptation has been studied to bridge domain gaps from different perspectives. One of the important tasks is cross-viewpoint domain adaptation \cite{rahmani2015learning,kong2017deeply,liu2017enhanced,sigurdsson2018actor,li2018unsupervised,zhu2021msnet,zhu2023leveraging}.
These works focused on learning geometric transformations of a camera but neglected other domain shifts such as environment differences. To learn viewpoint-invariant representations, 3D representations such as skeletons and human meshes are used as model inputs ~\cite{liu2017enhanced,sigurdsson2018actor}.
The other stream for video domain adaptation focuses on environmental changes.
Some of the recent works applied adversarial training for domain alignment \cite{jamaldeep,Chen_2019_ICCV,pan2019adversarial}. Specifically, Gradiant Reverse Layer ~\cite{li2018unsupervised} was adapted to C3D~\cite{tran2015learning}, TRN~\cite{Zhou_2018_ECCV} or both~\cite{pan2019adversarial} architectures. Chen \etal~\cite{Chen_2019_ICCV} proposed an attention-based model to attend to the temporal dynamics of videos. Pan \etal~\cite{pan2019adversarial} introduced a cross-domain attention model to learn relevant information. In contrast to previous literature which used complete videos as model inputs, we propose a model which leverages adversarially generated masks to better align the source and target domain.

\paragraph{Masked Visual Modeling.} Masked visual modeling has gained attention in both Natural Language Processing and Computer Vision. It learns effective representations by masking and reconstruction. Some early works\cite{VincentLLBM10} treated the masks as a noise type~\cite{VincentLBM08} or missing regions and used inpainting objectives~\cite{PathakKDDE16}.
More recently, transformer backbones became more and more popular due to their flexibility to mask different patches~\cite{beit,dong2021peco,he2021masked,wei2021masked,xie2021simmim,zhou2021ibot,zhu2023stmt}. BEiT~\cite{beit} followed the success of BERT~\cite{devlin2019bert} and proposed methods to learn visual representations by predicting the discrete tokens~\cite{dalle}. He~\cite{he2021masked} proposed an encoder-decoder architecture for masked image modeling. \cite{shi2022adversarial} proposed to adversarially learn the masks for masked visual modeling and performed evaluation on the image classification task. VideoMAE \cite{tong2022videomae} was inspired by MAE \cite{he2021masked} and adapted mask and reconstruction strategies to the video domain. One of the recent works applied the masked image modeling strategy for unsupervised domain adaptation on the image classification task \cite{hoyer2023mic}. 
In contrast to previous methods which use random masking strategies, we propose to generate challenging masks for the domain adaptation task in an adversarial way. Moreover, we do not use any reconstruction objectives and simply force the masked and unmasked views to have consistent predictions. 
\section{Method}
\begin{figure*}[t]
\centering
\makebox[\textwidth][c]{\includegraphics[width=1.0\textwidth, height=4.5cm]{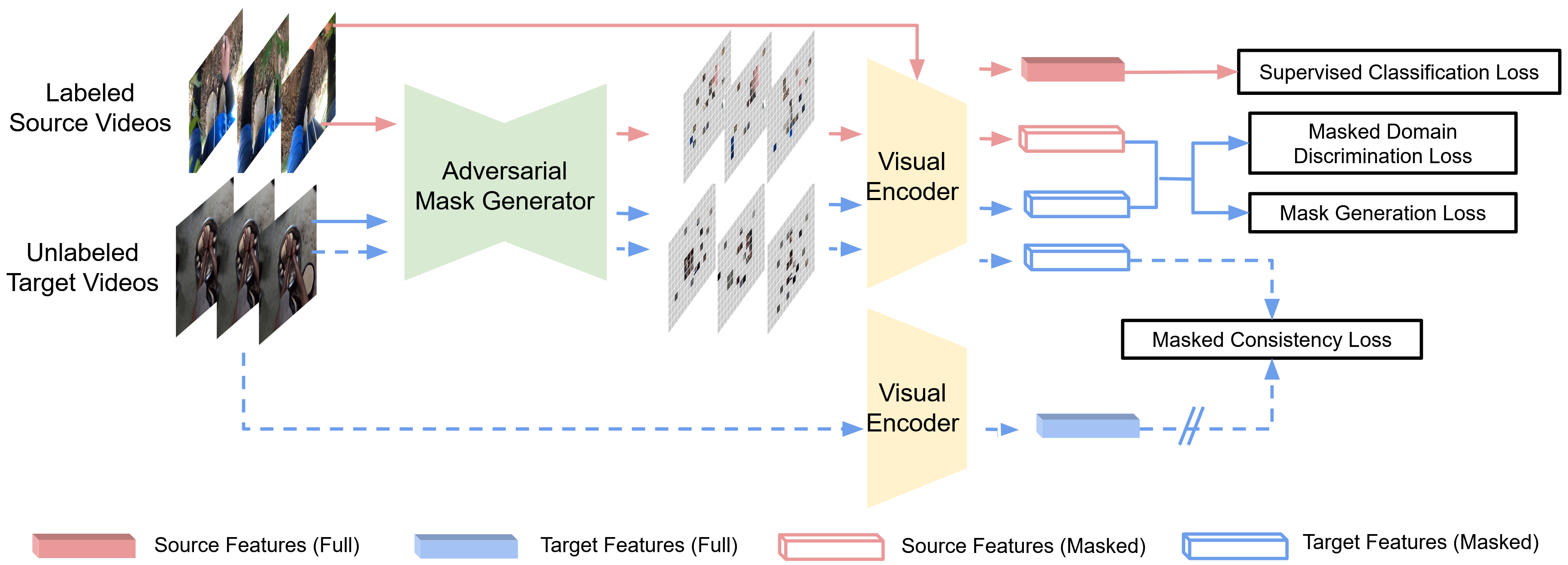}}%
\caption{Overview of the proposed framework. There are two training stages to learn domain-invariant and class-discriminative representations. The goal of stage one (denoted by solid lines) is to align the source and target domains. As directly aligning the two domains using full views may lead to trivial solutions, we propose an adversarial mask generator to produce masked samples. This module is trained with the domain-invariant encoder in an adversarial way. For the training of stage two (denoted by dashed lines), we propose a Masked Consistent Learning module to enhance the model's understanding of the spatial-temporal context, and thus increase the class-discrimination ability. We first initialize the class-discriminative visual encoder using weights learned in stage one. Then we force the visual encoder to have consistent predictions on the full and masked views of the same target video. Our two-stage training framework learns effective domain-invariant and class-discriminative representations, with robustness to large domain gaps. 
} 
\label{fig:method}
\vspace{-3mm}
\end{figure*}
\subsection{Overview}
In this section, we introduce our model for unsupervised video domain adaptation. Contrast to previous methods \cite{tzeng2017adversarial,
ganin2016domain,chen2019temporal,sahoo2021contrast,munro20multi} which take full data views as inputs to train a domain-alignment or class-discriminative loss, we aim at developing a model that can benefit from the masked forms for better domain alignment and context understanding. Following previous works for unsupervised domain adaptation \cite{sahoo2021contrast,tzeng2017adversarial,munro2020multi}, we adopt a multi-stage training schema. Our model consists of two stages. For stage one, We train the Generative Adversarial Domain Alignment Network. Specifically, the adversarial mask generator and domain-invariant visual encoder are trained in an adversarial way. The adversarial mask generator aims at producing challenging samples to maximize the distance between the source and target domain. Conversely, the domain-invariant visual encoder is trained with those masked samples to minimize the domain distance. For stage two, we further fine-tune the domain-invariant encoder by forcing the label predictions between the masked and full unlabeled videos to be consistent. As there are no ground truth labels for the target samples, we generate the pseudo-labels using complete videos. In summary, our proposed method has the following key components:
\begin{itemize}
    \item \textbf{Adversarial Mask Generator} is trained to generate challenging masks that will maximize the domain gap between masked source and target samples.
    \item \textbf{Domain-Invariant Visual Encoder} is trained to minimize the domain gap between the masked source and target samples. It is trained with Adversarial Mask Generator in an adversarial way.
    \item \textbf{Masked Consistency Loss} enforces the masked target videos and their full forms to have consistent label predictions.  
    \item \textbf{Class-Discriminative Visual Encoder} is initialized with the weights of Domain-Invariant Visual Encoder trained in stage one, and is fine-tuned using the masked consistency loss. 
\end{itemize}

See Figure~\ref{fig:method} for a high-level summary of the model, and
the sections below for more details.

\subsection{Unsupervised Domain Adaptation}
Given a set of labeled source videos $\mathcal{D}_{S}\!=\!\{(\mathbf{V}^{i\{s\}}, y^{i})\}_{i=1}^{N_{S}}$ and unlabelled target videos $\mathcal{D}_{T}\!=\!\{\mathbf{V}^{i\{t\}}\}_{i=1}^{N_{T}}$, the goal of UDA task is to learn a model $\mathcal{H}$ which minimizes the task risk $\epsilon_{\mathcal{D}_T}(\mathcal{H})$ in the target domain, $i.e. \ \epsilon_{\mathcal{D}_T}(\mathcal{H})=\mathbb{P}_{\mathcal{D}_T}[\mathcal{H}(x)\neq \mathcal{H}^*(x))]$, where $\mathcal{D}_T$ is the unlabeled target samples, and $\mathcal{H}^*$ is the ideal model in all model space. 
 The model $\mathcal{H}$ consists of a feature extractor $\mathcal{F}$ and a classification head $\mathcal{G}$, $i.e. \ {\mathcal{H}}(x) = \mathcal{G}(\mathcal{F}(x))$.

\subsection{Generative Adversarial Domain Alignment Network}
\label{sec: domain}
To align the source and target domains, adversarial adaptive learning methods \cite{tzeng2017adversarial,
ganin2016domain,munro20multi}  are used to regularize the source and target representations, so as to minimize the distance between the empirical source and target mapping distributions: $\mathcal{H}(x_s)$ and
$\mathcal{H}(x_t)$. Adversarial domain alignment is one of the commonly used strategies. A domain discriminator is trained along with the gradient reverse layer (GRL) \cite{ganin2016domain} to minimize the domain distance. The gradient reversal mechanism ensures that the distributions over the two domains are forced to be
similar (as indistinguishable as possible for the domain classifier), thus resulting in domain-invariant representations. Given an ideal model that is domain-invariant, the source classification
model can be directly applied to classify the samples from the target domain. Given a binary domain label, $d$, indicating if an example $x \in \textbf{S}$ or $x \in \textbf{T}$, the domain discriminator is defined as,
\begin{equation}
    \mathcal{L}_{d} = \sum_{x \in \{\textbf{S}, \textbf{T}\}} -d \log(\mathcal{D}(\mathcal{F}(x))) - 
    (1-d) \log(1-\mathcal{D}(\mathcal{F}(x)))
    \label{eq:advb}
\end{equation}

Where $\mathcal{F}$ is the visual encoder and $\mathcal{D}$ is the domain classification head. However, there are multiple factors that could lead to domain gaps, including different
backgrounds, lighting conditions, viewpoints, interacted objects, and motion variances. Existing state-of-the-art methods \cite{tzeng2017adversarial,ganin2016domain,chen2019temporal,sahoo2021contrast} directly use Eqn.~\ref{eq:advb} to align two domains based on the full input
data. However, it might lead to trivial solutions. To tackle this problem, we propose to minimize the distance between the empirical source and target mapping distributions learned from the \textbf{masked forms}: $h(m_s \odot x_s)$ and
$h(m_t \odot x_t)$, where $m_s$ and $m_t$ are element-wise masks and $\odot$ is the Hadamard product. The model can learn from samples that are adversarially masked based on domain classification loss.  The masked domain discrimination loss $\mathcal{L}^{m}_{d}$ is defined as:

\begin{equation}
\small
    \mathcal{L}^{m}_{d} = \sum_{x \in \{\textbf{S}, \textbf{T}\}} -d \log(\mathcal D(\mathcal F(x \odot m))) - 
    (1-d) \log(1-\mathcal D(\mathcal F(x\odot m)))
    \label{eq:adv}
\end{equation}

There are multiple video masking options, such as pixel-wise masking, tube masking, and frame-wise masking. Another important aspect is the masking ratio. He \emph{et. al} \cite{he2021masked} demonstrated that large mask ratios are essential for effective self-supervised learning. In contrast to previous methods, we propose to learn the mask in an adversarial way. The learned masks save the efforts to adjust masking hyper-parameters, and can produce challenging samples for domain-invariant learning. 
Given an RGB video $\x \in \RR^{t \times c \times w \times h}$, 
the adversarial mask generation model $\M$ produces an element-wise mask $\m=\M(\x)$, which is in the shape of $\RR^{t \times c \times w \times h}$ with values in $[0, 1]$. The generation model is trained with the objective of maximizing the distribution shifts between the two domains.
On the other hand, the domain-invariant  visual encoder  $\mathcal F$  takes the masked videos from the source and target domains
as inputs, and tries to minimize the mapping distributions.
The two models are jointly learned  using the following function:
\begin{align}
    \M^\star, \mathcal F^\star = \arg\min_\mathcal F \max_\M  \mathcal{D}^{m}(\x_s,\x_t; \mathcal{F}, \M)\,. \label{eq:minmax_obj} 
\end{align}

Where $\mathcal {D}^{m}$ denotes the masked feature distance. To stabilize the training process, inspired by Generative Adversarial Network (GAN) \cite{goodfellow2014generative}, we first freeze the mask generator and train the visual encoder using masked domain discrimination loss with GRL. In this way, the visual encoder learns domain-invariant representations that are as indistinguishable as possible for the domain classifier. Then we freeze the visual encoder and train the mask generator with masked domain discrimination loss only (without GRL). The mask generator learns to generate challenging masked videos which will maximize the domain distance (\emph{i.e}, the masked views of the source and target videos are as distinguishable as possible for the domain classifier). The adversarial mask generator 
$\M$ consists of a U-Net architecture \cite{ronneberger2015unet} and a pixel-wise softmax layer $\sigma$ to ensure that the sum of the generated mask equals one.

\subsection{Masked Consistency Learning}
The goal of the Generative Adversarial Domain Alignment Network is to learn domain-invariant representations using masked views. However, the models learned on labeled source videos may overfit the source domain but are less discriminative for the target \cite{9219132}. To help the model learn effective class-discriminative features, we enforce the model to make consistent predictions on the full and masked videos.

Specifically, we use an Adversarial Mask Generator trained in stage one to generate masked samples for unlabeled target videos. Moreover, we take the full video forms as inputs to generate pseudo-labels, and force the model to have consistent predictions on the masked and full videos. The proposed consistency learning module has two purposes: (1) Using the masked samples as a type of strong data augmentation. Based on \cite{zhang2022divide}, the unlabeled target samples can be divided into two groups: source-like samples and target-specific samples. The source-like samples can already be easily classified after the domain alignment; the target-specific samples, however, are more likely to confuse the video classification model. We aim to apply the adversarial mask generation model in Sec~\ref{sec: domain} to generate more target-like samples. 
(2) Enforcing the model to have consistent predictions for better context learning. To recognize a human action, a model can utilize clues from different parts of the video. This can be local spatial information, which originates from the same region as the corresponding cell in the feature map, or context information, which comes from nearby patches in the spatial-temporal domain that can belong to different parts of the object or its background~\cite{hoyer2023mic,hoyer2019grid}. 
The proposed Masked Consistency Learning (MCL) can help the model learn context relations on the unlabeled target domain, which will further improve the class-discriminative ability.

Specifically, MCL first generates adversarial masks using the mask generator trained in stage one, and then applies an element-wise multiplication of mask and video. In this way,
the masked target prediction $\hat{y}^{M}$ can only rely on the limited information of the remaining video pixels:
\begin{equation}
    \hat{y}^{M} = \mathcal{G}(\mathcal{F}(x\odot m))\,
\end{equation}
This makes the prediction more difficult. The masked  consistency loss $\mathcal{L}_c^m$ can be represented as
\begin{equation}
    \mathcal{L}_c^m = \sum_{x \in \{\textbf{T}\}}  -p^\mathit{T} \log \hat{y}^{M}
\end{equation}
where $p^T$ denotes a pseudo-label. 
The proposed model uses pseudo-labels as there is no ground truth available for the target domain. The pseudo-label is the prediction of the visual encoder and classification head of the complete target video $x$. 
\begin{equation}
    p^\mathit{T} = \argmax \mathcal{G}(\mathcal{F}(x))\,.
    \label{eq:pseudo_label}
\end{equation}

\subsection{Training}
The model training includes two stages. For the first stage, we train the domain-invariant visual encoder using the masked domain discrimination loss $\mathcal{L}^{m}_{d}$ and the supervised classification loss $\mathcal{L}_\mathcal{S} $. Training with supervised classification loss expects the presence of labels and thus can only be applied to the labeled source input. The supervised classification loss can be represented as: $\mathcal{L}_\mathcal{S} = \sum_{i=1}^N \text{CE}({cls}^i,{cls}^{\ast i})$, where ${cls}^i$ is the predicted label and ${cls}^{\ast i}$ is the ground truth label, and $N$ is the number of labeled source videos. To stabilize the training, the mask generator and visual encoder training proceed in alternating periods. In stage two, we freeze the mask generator and train the visual backbone using the masked consistency loss $\mathcal{L}^{m}_{c}$. 

\section{Experiment}

\begin{figure*}[t]
\centering
\makebox[\textwidth][c]{\includegraphics[width=1.0\textwidth, height=4.2cm]{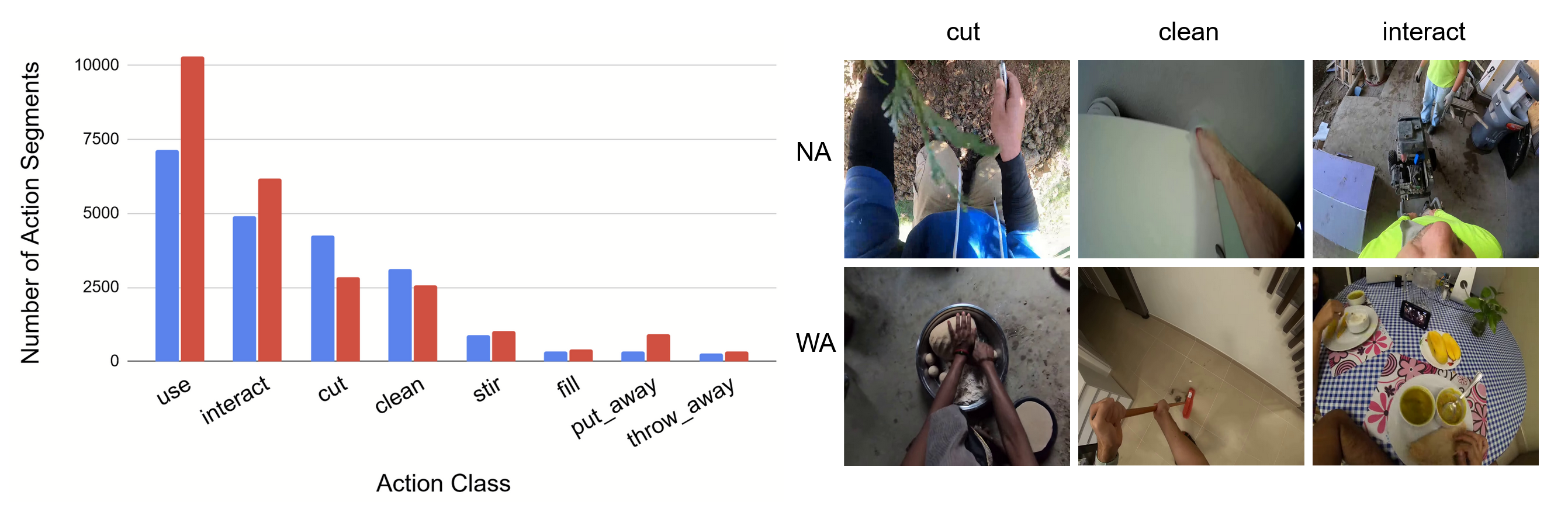}}%
\caption{\textbf{Left:} Class distributions per domain for the U-Ego4D benchmark. \textbf{Right:} Videos collected from different regions are treated as different domains. Different from the Epic-Kitchen dataset which is limited to the kitchen scenario, the same action in the U-Ego4D benchmark can happen in totally different environments.
}

\label{data}
\end{figure*}

\subsection{Datasets}
\paragraph{Epic-Kitchen.} We evaluate our model on the commonly-used Epic-Kitchen dataset \cite{Damen2018EPICKITCHENS}. Epic-Kitchen contains egocentric videos for fine-grained activities in the kitchen environment. It has three domains (D1, D2, and D3), and each domain is a different kitchen. The task is to adapt between
each pair of kitchens, which have different visual appearances. This benchmark contains 8
verb action classes, which occur in combination with different nouns. We use the
standard training and test splits provided by the previous works \cite{Damen2018EPICKITCHENS, sahoo2021contrast} to conduct our experiments.
\paragraph{U-Ego4D.} We construct a new unsupervised domain adaptation benchmark called U-Ego4D. It builds on the massive-scale Ego4D dataset  \cite{Ego4D2022CVPR}, which records daily-life activity videos spanning hundreds of scenarios (\emph{e.g.} household, outdoor, workplace, leisure). 
We select two regions with the largest number of videos, North America and West Asia as two different domains (\emph{i.e.} domain NA and domain WA).   In this way, we can analyze the domain gaps between different regions. Moreover, in U-Ego4D, the same action can happen in the same or different scenarios (\emph{e.g.} indoors and outdoors), which increases the action diversities. For example, in Figure \ref{data}, the action "interact" can happen indoors (during a meal) or happen outdoors. For the action categories, we select the 8 largest classes: (use, interact, clean, put\_away, cut, throw\_away, stir, and fill). The class distributions are shown in Figure  \ref{data}. The in-balanced class distributions bring additional challenges for the domain adaptation task. Besides, U-Ego4D is \textbf{3x} larger than Epic-Kitchen in terms of video length and clip number. Specifically, U-Ego4D has 35.67 hours of video with 35,937 video clips in total, while Epic-Kitchen only has 7.98 hours with 10,094 clips. Please refer to supplementary materials for data pre-processing details.

\subsection{Implementation Details} We use the transformer-based architecture, ViT-Small\cite{tong2022videomae},  as our visual encoder. Following previous works \cite{tzeng2017adversarial,
ganin2016domain,
chen2019temporal,
sahoo2021contrast}, the model is initialized with Kinetics-400 \cite{kay2017kinetics} pre-trained weights. Our model uses the same visual encoder and initialization strategies as the baselines with ViT backbones for a fair comparison. Following previous works for unsupervised domain adaptation \cite{sahoo2021contrast,tzeng2017adversarial,munro2020multi}, we adopt a multi-stage training schema. The adversarial mask generator and domain-invariant encoder are trained in stage one. %and the masked consistency loss and class-discriminative visual encoder are trained in stage two. 
The class-discriminative visual encoder is trained using the masked consistency loss in stage two. 
We use a clip length of 16 with a spatial size of 224 $\times$ 224 to train all the models. AdamW \cite{loshchilov2018decoupled} is used with $\beta$ equals (0.9, 0.999) as the model optimizer with a learning rate 1e-4.  We use batch size 8 for all the experiments. For the implementation of the mask generator, we use a U-Net structure with a depth of 4. For inference, we use 16 randomly
sampled frames per video and use the visual encoder along with the classification head to recognize the action. For more details, please see the supplementary material. 

\subsection{Baselines} We compare our model with state-of-the-art unsupervised domain adaptation models. ADDA \cite{tzeng2017adversarial} is a general framework which combined discriminative modeling, untied weight sharing, and a GAN loss for unsupervised domain adaptation. DANN~\cite{ganin2016domain} proposed gradient reverse layer (GRL) for domain-invariant representation learning. TA\textsuperscript{$3$}N~\cite{chen2019temporal} adapted attention mechanisms to explicitly attend to the temporal dynamics using domain discrepancy for effective domain alignment. CoMix is a contrastive learning framework which leveraged background mixing to produce augmented samples \cite{sahoo2021contrast}. TransVAE \cite{wei2023unsupervised} combines seven different loss functions \cite{chen2019temporal,ganin2016domain,Zhou_2018_ECCV} for spatial-temporal disentanglement and domain gap minimization. All of those models were originally implemented with the I3D backbone. We also run experiments for DANN~\cite{ganin2016domain} and CoMix \cite{sahoo2021contrast} by replacing I3D with the same ViT backbone as our proposed model for a fair comparison. Note that it is not feasible to replace the backbone of TransVAE  with ViT, as some of the loss functions in TransVAE are specifically designed for its architecture. All of these models use single modality features as our proposed method. There are several recent works conducting video-based unsupervised domain adaptation using multi-modal data which combines RGB and Flow. Although our method solely uses RGB
information, we still take this set of methods into account following the previous work \cite{wei2023unsupervised}. Specifically, we consider MM-SADA \cite{munro2020multi},
STCDA \cite{song2021spatio}, CMCD \cite{sahoo2021contrast}, CleanAdapt \cite{dasgupta2022overcoming}, MixDANN \cite{yin2022mix} and CIA \cite{yang2022interact}.

\begin{table*}[t]
\footnotesize
\begin{center}
\resizebox{\linewidth}{!}{
\begin{tabular}{l | c | cccccc | c}
\Xhline{2\arrayrulewidth}  
\multirow{2}{*}{\textbf{Method}} & \multirow{2}{*}{\textbf{Backbone}}  & \multicolumn{6}{c|}{\textbf{Epic-Kitchens}} & \multirow{2}{*}{\textbf{Average}}\\ 
\cline{3-8}
&  & \textbf{D2$\rightarrow$D1} & \textbf{D3$\rightarrow$D1} & \textbf{D1$\rightarrow$D2} & \textbf{D3$\rightarrow$D2} & \textbf{D1$\rightarrow$D3} & \textbf{D2$\rightarrow$D3} & \\
\Xhline{1\arrayrulewidth}  
Supervised Source        & I3D          & $35.4$    & $34.6$        & $32.8$        & $35.8$        & $34.1$        & $39.1$        & $35.3$   \\

ADDA~\cite{tzeng2017adversarial}                 & I3D         & $36.3$    & $36.1$        & $35.4$        & $41.4$        & $34.9$        & $40.8$        & $37.4$   \\
DANN~\cite{ganin2016domain}                  & I3D        & $38.3$    & $38.8$        & $37.7$        & $42.1$        & $36.6$        & $41.9$        & $39.2$  \\
TA\textsuperscript{$3$}N~\cite{chen2019temporal}    & I3D       & $40.9$    & $39.9$        & $34.2$        & $44.2$        & $37.4$        & $42.8$        & $39.9$   \\
CoMix \cite{sahoo2021contrast} & I3D            & $38.6$ & $42.3$ & $42.9$ & $49.2$ & $40.9$ & $45.2$ & $43.2$ \\
% \Xhline{1\arrayrulewidth}  
TranSVAE \cite{wei2023unsupervised} & I3D            & $50.3$ & $48.0$ & $50.5$ & $58.0$ & $50.3$ & $\textbf{58.6}$ & $52.6$ \\
\Xhline{1\arrayrulewidth}  
\cellcolor[gray]{0.8} Supervised Target &\cellcolor[gray]{0.8} I3D            &\cellcolor[gray]{0.8} $57.0$ &\cellcolor[gray]{0.8} $57.0$ &\cellcolor[gray]{0.8} $64.0$ &\cellcolor[gray]{0.8} $64.0$ &\cellcolor[gray]{0.8} $63.7$ &\cellcolor[gray]{0.8} $63.7$ &\cellcolor[gray]{0.8} $61.5$ \\
\Xhline{1\arrayrulewidth}

Supervised Source        & ViT          & $44.7$    & $45.6$        & $53.3$        & $55.6$        & $46.7$        & $47.8$        & $49.0$   \\
DANN~\cite{ganin2016domain}                  & ViT        & $49.8$    & $47.5$        & $58.9$        & $57.3$        & $53.8$        & $52.4$        & $53.3$  \\

% \Xhline{1\arrayrulewidth}  
CoMix \cite{sahoo2021contrast} & ViT            & $46.3$ & $47.3$ & $56.7$ & $59.3$ & $51.4$ & $52.3$ & $52.2$ \\
\textbf{Ours} & ViT           & $\textbf{50.7}$ & $\textbf{48.2}$ & $\textbf{64.6}$ & $\textbf{60.8}$ & $\textbf{55.5}$ & $56.6$ & $\textbf{56.1}$ \\
\Xhline{1\arrayrulewidth}  
\cellcolor[gray]{0.8} Supervised Target & \cellcolor[gray]{0.8} ViT            &\cellcolor[gray]{0.8} $57.6$ &\cellcolor[gray]{0.8} $57.6$ &\cellcolor[gray]{0.8} $66.5$ &\cellcolor[gray]{0.8} $66.5$ &\cellcolor[gray]{0.8} $67.2$ &\cellcolor[gray]{0.8} $67.2$ &\cellcolor[gray]{0.8} $63.8$ \\

\Xhline{1\arrayrulewidth} 
\Xhline{2\arrayrulewidth}  
\end{tabular}}

\end{center}
\caption{ Experimental Results on Epic-Kitchens Dataset. Our model achieves the best average performance among all state-of-the-art methods.} 
\label{epic_main} 

\end{table*}

\begin{table*}[t]
\footnotesize
\begin{center}
\resizebox{\linewidth}{!}{
\begin{tabular}{l | cccccc | c}
\Xhline{2\arrayrulewidth}  
\multirow{2}{*}{\textbf{Method}}  & \multicolumn{6}{c|}{\textbf{Epic-Kitchens}} & \multirow{2}{*}{\textbf{Average}}\\ 
\cline{2-7}
&   \textbf{D2$\rightarrow$D1} & \textbf{D3$\rightarrow$D1} & \textbf{D1$\rightarrow$D2} & \textbf{D3$\rightarrow$D2} & \textbf{D1$\rightarrow$D3} & \textbf{D2$\rightarrow$D3} & \\
\Xhline{1\arrayrulewidth}  
Supervised Source                & $43.0$    & $43.0$        & $43.2$        & $55.5$        & $42.5$        & $48.0$        & $45.9$   \\

MM-SADA~\cite{munro2020multi}                          & $48.2$    & $50.9$        & $49.5$        & $56.1$        & $44.1$        & $52.7$        & $50.3$   \\
STCDA~\cite{song2021spatio}                        & $49.0$    & $\textbf{52.6}$        & $52.0$        & $55.6$        & $45.5$        & $52.5$        & $51.2$  \\
CMCD~\cite{sahoo2021contrast}         & $49.5$    & $48.7$        & $50.3$        & $56.3$        & $46.3$        & $52.0$        & $51.0$   \\
% \Xhline{1\arrayrulewidth}  
% \Xhline{1\arrayrulewidth}  
CleanAdapt \cite{dasgupta2022overcoming}             & $46.2$ & $47.8$ & $52.7$ & $54.4$ & $47.0$ & $52.7$ & $50.3$ \\
MixDANN \cite{yin2022mix}             & $50.3$ & $51.0$ & $56.0$ & $54.7$ & $47.3$ & $52.4$ & $52.0$ \\
CIA \cite{yang2022interact}             & $49.8$ & $52.2$ & $52.5$ & $57.6$ & $47.8$ & $53.2 $ & $52.2$ \\

\textbf{Ours}           & $\textbf{50.7}$ & $48.2$ & $\textbf{64.6}$ & $\textbf{60.8}$ & $\textbf{55.5}$ & $\textbf{56.6}$ & $\textbf{56.1}$ \\
\Xhline{1\arrayrulewidth}

\Xhline{1\arrayrulewidth} 
\Xhline{2\arrayrulewidth}  
\end{tabular}}

\end{center}
\caption{ Experimental Results on Epic-Kitchen with comparisons to approaches using multi-modality data as the input. Our model, which only uses RGB videos, achieves the best average performance among all state-of-the-art methods.} 
\label{epic_main_multi} 
\vspace{-5mm}
\end{table*}

\begin{table*}[t]
\scriptsize
\begin{center}

\resizebox{0.8 \linewidth}{!}{
\begin{tabular}{l | c | cc | c}
\Xhline{1\arrayrulewidth}  
\multirow{2}{*}{\textbf{Method}} & \multirow{2}{*}{\textbf{Backbone}}  & \multicolumn{2}{c|}{\textbf{U-Ego4D}} & \multirow{2}{*}{\textbf{Average}}\\ 
\cline{3-4}
&  & \textbf{NA$\rightarrow$WA} & \textbf{WA$\rightarrow$NA} &  \\ 

\Xhline{1\arrayrulewidth}
Supervised Source    & ViT  &  52.5& 48.1&50.3\\
DANN \cite{ganin2016domain}      & ViT    & $56.9$    & $48.6$         & $52.8$     \\
CoMix \cite{sahoo2021contrast}                & ViT        & $56.3$    &$48.9$          & $52.6$  \\
\textbf{Ours}   & ViT           & $\textbf{58.4}$ & $\textbf{49.4}$  & $\textbf{53.9}$  \\
\Xhline{1\arrayrulewidth} 
\cellcolor[gray]{0.8} Supervised Target   & \cellcolor[gray]{0.8} ViT    &\cellcolor[gray]{0.8} 67.5  &\cellcolor[gray]{0.8} 71.3&\cellcolor[gray]{0.8} 69.4\\

\Xhline{1\arrayrulewidth}  
\end{tabular}}

\vspace{-3mm}
\end{center}
\caption{ Experimental Results on the proposed U-Ego4D Dataset. Different from Epic-Kitchen which specifically focuses on domain transfer among different kitchens, U-Ego4D focuses on a more practical setting: domain adaptation between different regions. Our model achieves the best performance compared to state-of-the-art models.} 
\label{ego_main}
\end{table*}

\subsection{Main Results}
\paragraph{Results on Epic-Kitchen.} We compare our model with state-of-the-art unsupervised video domain adaptation models on the Epic-Kitchen dataset and report the results in Table \ref{epic_main}. Our model achieves the best performance on the 5 of 6 splits as well as the
best average performance of 56.1\%.
Besides, we observe that the performance gap between our method and the supervised target model has been reduced to 7.7\%, which demonstrates the potential to close the domain gap using masked video modeling methods. 

\paragraph{Compare to Multi-Modal Methods.} We further compare our model with recent video-based unsupervised domain adaptation methods
that use multi-modalities, \emph{i.e.}, RGB features and optical flows, although our model only uses RGB
features. The results are reported in Table~\ref{epic_main_multi}. We observe that our method achieves the best average result among all of the six multi-modal
methods.

\paragraph{Results on U-Ego4D.} We compare our model with state-of-the-art unsupervised video domain adaptation models on the proposed U-Ego4D benchmark and report the results in Table~\ref{ego_main}. Our model achieves the best performance including the
best average performance of 53.9\%. Although our model achieves promising performance, there is still a large performance gap between our method and the supervised target model, which is 15.5\%. This shows the great potential for
further improvement to bridge the large domain gap caused by different regions. Besides, we observe that on Epic-Kitchen, all the methods can help increase the performance from Supervised-Source by at least an absolute 3.2\% (CoMix) to 7.1\% (ours), while on U-Ego4D we are only seeing an increase of 2.5\% (CoMix) to 3.6\% (ours). Low-bound (Supervised-Source) methods on both datasets have similar performance (49.0\% v.s. 50.3\%), while U-Ego4D has a higher upper bound (supervised-target) 69.4\% compared to Epic-Kitchen which is 63.8\%. This suggests that there is still a larger domain gap after applying the state-of-the-art domain adaptation methods and thus the proposed U-Ego4D dataset is more challenging.
\begin{table*}[t]
\begin{center}
\resizebox{1\linewidth}{!}{
\begin{tabular}{l | c | cccccc | c}
\Xhline{2\arrayrulewidth}  
\multirow{2}{*}{\textbf{Method}} & \multirow{2}{*}{\textbf{Backbone}}  & \multicolumn{6}{c|}{\textbf{Epic-Kitchen}} & \multirow{2}{*}{\textbf{Average}}\\ 
\cline{3-8}
&  & \textbf{D2$\rightarrow$D1} & \textbf{D3$\rightarrow$D1} & \textbf{D1$\rightarrow$D2} & \textbf{D3$\rightarrow$D2} & \textbf{D1$\rightarrow$D3} & \textbf{D2$\rightarrow$D3} & \\
\Xhline{1\arrayrulewidth}  

\Xhline{1\arrayrulewidth}
S.O.        & ViT          & $44.7$    & $45.6$        & $53.3$        & $55.59$        & $46.7$        & $47.8$        & $49.0$     \\
S.O.+GADAN                  & ViT        & $50.2$    &$47.9$&   $61.0$        & $56.8$        & $54.3$        & $53.3$               & $53.9$  \\
S.O.+GADAN+MCL   & ViT           & $\textbf{50.7}$ & $\textbf{48.2}$ & $\textbf{64.6}$ & $\textbf{60.8}$ & $\textbf{55.5}$ & $\textbf{56.6}$ & $\textbf{56.1}$  \\
\Xhline{1\arrayrulewidth} 
\Xhline{2\arrayrulewidth}  
\end{tabular}}

\end{center}
\caption{Ablation Studies on Epic-Kitchen. Each of our proposed modules brings stable performance improvement in all tasks. S.O. stands for the source-only model, GADAN stands for Generative Adversarial Domain Alignment Network, and MCL stands for Masked Consistency Learning.}
\label{oursaba} 
\end{table*}

\subsection{Ablation Studies and Analysis}
\paragraph{Effectiveness of Different Components.} We test various ablations of
our model on the Epic-Kitchen dataset to substantiate our design decisions. The results are shown in Table~\ref{oursaba}. We observe that each component
of our model brings consistent improvements in all six splits. Overall, compared to the source-only baseline, Generative Adversarial Domain Alignment Network improves the average accuracy from 49.0\% to 53.9\%, and Masked Consistency Learning can further improve the accuracy to 56.1\%.

\begin{SCtable}
\footnotesize
\begin{tabular}{c|c}
\toprule
Method & Top-1  (\%)\\
\midrule
S.O.&  53.3\\ 
S.O.+DL &  58.9 \\ 
S.O.+MDL + random tube  (r=0.5)&59.8\\
S.O.+MDL + random tube  (r=0.75)&59.6\\
S.O.+MDL + random tube (r=0.9)&59.3\\
S.O. + AMG (GADAN)&\textbf{61.0}\\
% &\checkmark & 64.13 \\
% \checkmark&\checkmark & \textbf{65.59} \\
\bottomrule
\end{tabular}
\caption{Comparison of Different Loss Functions and Masking Strategies. 
% Supv stands for the supervised source-only baseline,
DL stands for the domain discrimination loss (with full-view inputs). MDL stands for the masked domain discrimination loss. AMG stands for our  adversarial mask generator.}
\label{tab:s1}
\end{SCtable}

\begin{SCtable}
\footnotesize
\begin{tabular}{c|c}
\toprule

Method& Top-1 (\%) \\ \midrule
GADAN& 61.0\\
GADAN + naive pseudo labeling& 62.2 \\
GADAN + $\mathrm {MCL}_\mathrm{CE}$ + random tube & 63.4 \\
GADAN + $\mathrm {MCL}_\mathrm{MSE}$ + AMG  & 62.8 \\
GADAN + $\mathrm {MCL}_\mathrm{CE}$ + AMG  & \textbf{64.6} \\
\bottomrule
\end{tabular}
\caption{Comparison of Different Loss Functions and Masking Strategies for MCL. CE and MSE stand for the cross entropy loss and mean squared error loss, respectively. AMG stands for our adversarial mask generator.}
\label{tab:s2}
\end{SCtable}

\paragraph{Comparison of Different Loss Functions and Masking
Strategies for GADAN.}
We test our Generative Adversarial Domain Alignment Network with different loss functions and masking strategies. The results are shown in Table~\ref{tab:s1}. We perform all the experiments with the same hyper-parameters on the D1$\rightarrow$D2 split of Epic-Kitchen for a fair comparison. Row 1 shows the performance of the source-only baseline.  Row 2 shows the performance of the source-only baseline with the domain discrimination loss (DL) and gradient reverse layer. For rows 3-5, we replace the masks produced by the Adversarial Mask Generator in GADAN with random tube masks. We test three mask ratios: 0.5, 0.75 and 0.9. The last row is the result of our full GADAN model. We observe that the proposed adversarial mask generator can bring 1.2\% performance improvement compared to the best model that is trained with random masks. Moreover, as our masks are directly learned using the mask generation objective, it saves the efforts to adjust the hyper-parameters, such as mask types and mask ratios.

\paragraph{Comparison of Different Loss Functions and Masking
Strategies for MCL.} We test MCL with different consistency losses and masking strategies. We perform all the experiments with the same hyper-parameters on Epic-Kitchen's D1$\rightarrow$D2 split. The results are shown in Table~\ref{tab:s2}. Row 1 shows the performance of GADAN. Row 2 shows the performance of GADAN with the naive pseudo-labeling method. Row 3 shows the performance of GADAN with MCL. Here we replace masks produced by AMG with random tubes. Comparing row 3 and row 5 (our full model), we find that adversarially generated masks lead to better performance for masked consistency learning. We also test different consistency losses including the cross-entropy loss (MCL\textsubscript{CE}) and mean squared error loss (MCL\textsubscript{MSE} ). For the cross-entropy loss, we force the hard labels predicted from masked and full views to be consistent. For the mean squared error loss, we force the soft logits of masked and full views to be consistent. Comparing row 4 and row 5, we find that cross-entropy loss leads to better performance compared to mean squared error loss.

\begin{figure*}[t]
\centering
\makebox[\textwidth][c]{\includegraphics[width=1.05\textwidth, height=6cm]{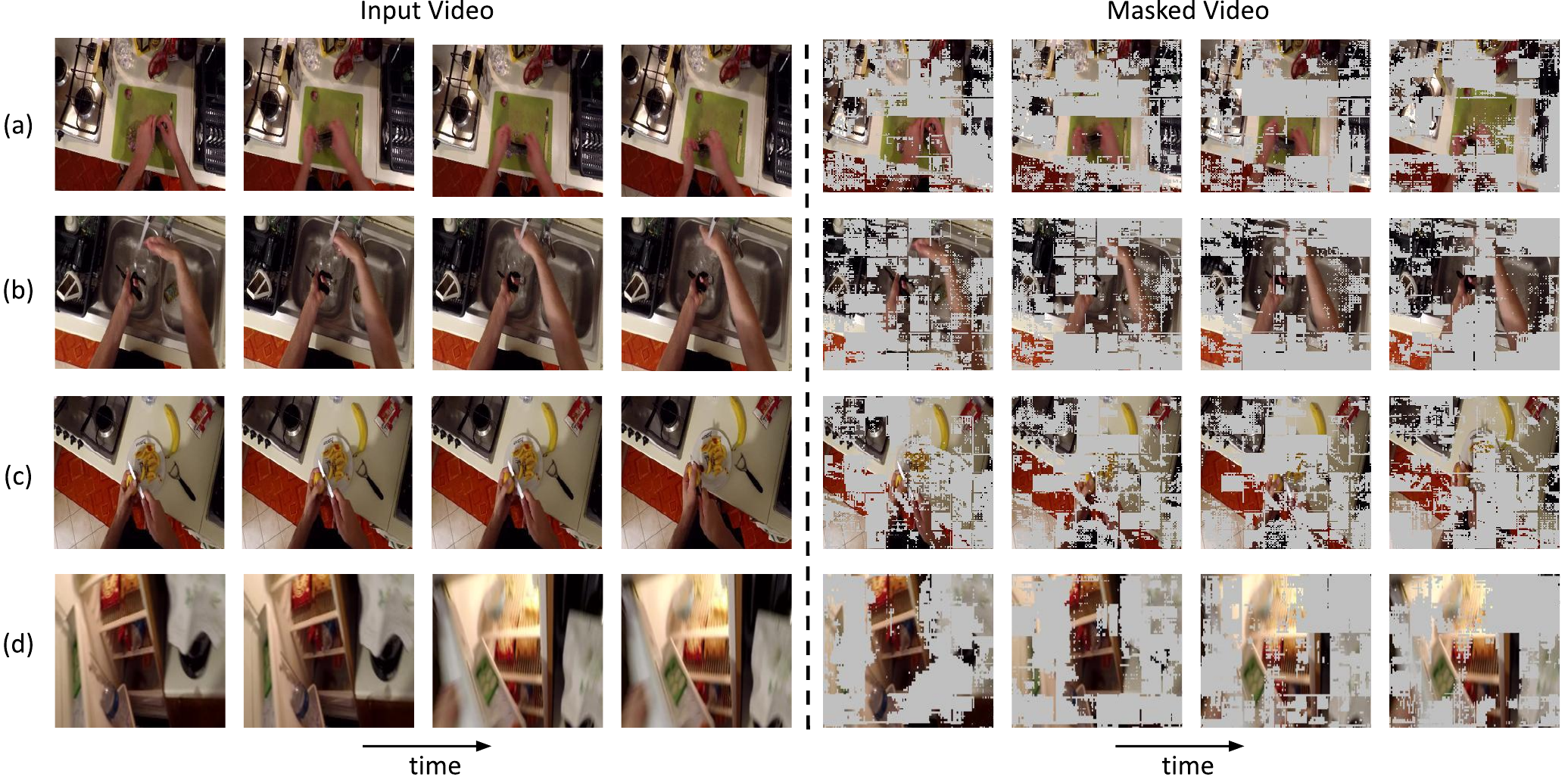}}%
\caption{ Visualizations of the Adversarially-Learned Masks.
}

\label{fig:maskvis}
\vspace{-3mm}
\end{figure*}

\paragraph{Visualizations of the Adversarially-Learned Masks.} We visualize the adversarially-learned masks in Figure~\ref{fig:maskvis}. We observe that after applying the learned masks, only the key instances are kept for each frame. Specifically, in (a) and (b), the regions that describe human-object interaction (person's hands, green board, sink) are retained. In (c) and (d), some of the key objects are preserved, such as bananas and refrigerators. In this way, the model is able to use the generated hard samples for domain-invariant and class-discriminative feature learning, and thus further improve performance.

\vspace{-3mm}

\subsection{Discussion} 
Although our model is trained and evaluated on eight different splits, 
potential dataset biases can still cause
negative societal impact in a real-world deployment. 
For example, due to the small size of the action datasets, they may not properly represent actions performed
by minority groups. Therefore, the models trained on these datasets (whether domain-adapted or not) might still under-represent some groups of people in the real world applications.

\section{Conclusion}
We have presented a novel transformer-based model 
for unsupervised domain adaptation in egocentric videos. 
We are the first to show that masked video modeling can benefit both domain-invariant and class-discriminative  feature learning. Our method also establishes new state-of-the-art performance on Epic-Kitchen and U-Ego4D. 
We believe our
dataset, together with our models, will facilitate future research in the domain adaptation and generalization field.

\bibliographystyle{splncs04}
\bibliography{main}
\end{document}